\documentclass[sigconf,natbib=false]{acmart}
\AtBeginDocument{%
  }

    \RequirePackage[
  datamodel=acmdatamodel,
  style=acmnumeric,
  ]{biblatex}

\begin{document}

\title{LinkedIn Profile Characteristics and Professional Success Indicators}


\author{Tania-Amanda Fredrick Eneye}
\affiliation{%
  \institution{Texas Tech University }
  \city{Lubbock}
  \state{Texas}
  \country{USA}
}
\email{tafredri@ttu.edu}

\author{Ashlesha Malla}
\affiliation{%
  \institution{Texas Tech University}
  \city{Lubbock}
  \state{Texas}
  \country{USA}}
\email{asmalla@ttu.edu}

\author{Pawan Paudel}
\affiliation{%
  \institution{Texas Tech University}
  \city{Lubbock}
  \state{Texas}
  \country{USA}}
\email{papaudel@ttu.edu}



\begin{abstract}
This study explores the relationship between LinkedIn profile characteristics and professional success, focusing on the indicators of promotions, follower count, and career progression rate. By leveraging a dataset of over 62,000 anonymized LinkedIn profiles, we developed predictive models using machine learning techniques to identify the most influential factors driving professional success. Results indicate that while promotions are highly predictable, follower growth exhibits greater complexity. This research provides actionable insights for professionals seeking to optimize their LinkedIn presence and career strategies.
\end{abstract}

\keywords{LinkedIn, Machine Learning, Professional Success, Profile Optimization, Career Progression}


\maketitle
 \section{Introduction}
LinkedIn has become an indispensable tool for professionals across the globe. Whether it's connecting with colleagues, searching for jobs, or building an online presence, the platform has transformed the way we approach career development. With over one billion users, LinkedIn offers countless opportunities to network, showcase skills, and grow professionally \cite{brochet2012linkedin}. Research has shown that professional social networking platforms significantly influence career development and job search outcomes \cite{utz2016linkedin}.
While there are countless tips and guides available, most lack a strong data-driven foundation, leaving professionals guessing about what will actually drive success. Previous studies have demonstrated that social capital accumulated through professional networks directly impacts career success, yet the specific characteristics of effective online professional profiles remain understudied.

Our project set out to answer an important question: how do LinkedIn profile characteristics and networking metrics correlate with professional success? For this study, we defined success using three measurable outcomes: promotions, follower count, and career progression rate. These metrics align with established research on career success indicators in the digital age, capturing both traditional indicators of career growth and modern aspects of professional presence.

We worked with a dataset of over 62,000 anonymized LinkedIn profiles, which included a diverse range of features \cite{ashish2023linkedin}. These features ranged from traditional career metrics like years of experience and promotion history to softer attributes like profile photo quality, represented as a beauty score. This approach builds upon previous work examining the role of visual elements in professional success \cite{edwards2015social} and the growing importance of digital self-presentation in career advancement \cite{roulin2019linkedin}.

To achieve this, we used advanced machine learning models, including Random Forest \cite{breiman2001random}, Gradient Boosting \cite{friedman2001greedy}, and CatBoost \cite{prokhorenkova2018catboost}, to predict professional success indicators. Each model offered unique insights, helping us identify which profile features mattered most for each success metric. This methodology follows established practices in analyzing social network data \cite{guy2011social} and career trajectory prediction.

This project goes beyond simply identifying what makes a LinkedIn profile "good." It offers a glimpse into how the professional landscape is evolving in the digital age, building on research about the transformation of professional networking \cite{barbaroux2016knowledge}. For individuals, it provides actionable recommendations for enhancing their profiles. For organizations and recruiters, it offers a deeper understanding of how LinkedIn metrics might align with indicators of professional potential \cite{zide2014linkedin}. More broadly, this study contributes to the growing body of literature on digital professional identity \cite{papacharissi2009virtual} and its impact on career outcomes.
 \section{Related Work and Background}
\subsection{Evolution of Professional Networking}
LinkedIn has become a vital platform for professional networking and career advancement, offering individuals a space to showcase their skills, connect with peers, and engage with industry leaders \cite{brochet2012linkedin}. It has redefined how professionals present themselves, blending traditional career elements like education and experience with modern factors such as social proof and networking visibility. However, despite its widespread use, the factors that truly contribute to success on LinkedIn remain unclear.
\subsection{Changing Paradigms of Career Success}
Historically, career success was measured through milestones like promotions or job titles. While these remain important, modern platforms like LinkedIn emphasize additional metrics such as follower count, engagement rates, and profile aesthetics. For example, a well-chosen profile picture or a clear, engaging summary may significantly impact a user's ability to attract opportunities. This shift reflects the growing complexity of professional success in the digital age, where visibility and influence often play as critical a role as experience.
\subsection{Profile Feature Analysis and Personality Perception}
Van de Ven et al. \cite{van2017personality} conducted pioneering work in analyzing how LinkedIn profiles reflect and communicate professional personality traits. Their study examined how recruiters and HR professionals interpret various profile elements to form impressions of candidates' personalities. This research is particularly relevant as it established a framework for quantifying subjective profile characteristics, supporting our investigation into how profile aesthetics influence professional success.
\subsection{Skills Analysis and Professional Identity}
Bastian et al. \cite{bastian2014linkedin} developed methods for large-scale skills extraction and inference from LinkedIn profiles. Their work demonstrated how machine learning techniques could be applied to understand and categorize professional skills across millions of profiles. While existing research on LinkedIn has primarily focused on recruitment trends or labor market insights, leaving a gap in understanding specific profile characteristics that drive success, Bastian's work provides a foundation for analyzing structured profile data at scale.
\subsection{Recruitment and Selection Methods}
Roulin and Levashina \cite{roulin2019linkedin} examined LinkedIn's potential as a selection method in recruitment, focusing on its psychometric properties and assessment capabilities. Their research validated LinkedIn as a legitimate tool for professional evaluation. This is particularly relevant given that popular advice for LinkedIn optimization often relies on anecdotal evidence rather than rigorous data analysis.
\subsection{Research Gap and Our Contribution}
While these studies have made valuable contributions to understanding LinkedIn profiles from various perspectives, there remains a gap in quantitatively connecting profile characteristics to measurable success indicators. Our research bridges this gap by combining multiple aspects of profile analysis into a unified predictive framework, introducing new metrics for measuring professional success, applying modern machine learning techniques to understand feature importance, and analyzing a larger, more diverse dataset than previous studies. This approach allows us to not only build upon existing research but also provide new insights into how specific profile characteristics correlate with career advancement and professional network growth.
 \section{Methodology}
\subsection{Data Collection}
The dataset used in this project was sourced from Kaggle and consists of 62,706 anonymized LinkedIn profiles\cite{ashish2023linkedin}. This dataset offers a diverse range of features, including professional metrics such as position length, promotions, and follower count, as well as profile characteristics like beauty scores, demographic details, and career progression indicators. These features were chosen to capture a holistic view of what influences professional success on LinkedIn.

The data was collected under ethical guidelines, ensuring that all personal identifiers were removed to protect user privacy. The scale and diversity of the dataset make it particularly suitable for exploring correlations and building predictive models. By including both traditional metrics (e.g., years of experience) and modern indicators (e.g., profile aesthetics), the dataset provides a robust foundation for understanding how LinkedIn profiles influence career trajectories. This comprehensive approach enables us to capture both the tangible and intangible aspects of professional success.

\subsection{Data Pre-processing and Feature Engineering}
Before analysis, the dataset underwent extensive cleaning and pre-processing to ensure its suitability for machine learning models. Missing values, a common challenge in large datasets, were addressed using median imputation for numeric variables and mode imputation for categorical variables. These techniques allowed us to preserve the integrity of the dataset without discarding valuable records. The 'glass' column presented a particular challenge with 47,314 missing values out of 62,706 total records (75.5\% missing). We addressed this through a two-step approach: creating a binary indicator for missing values, then applying mode imputation for the remaining cases.

Outliers posed another significant challenge, particularly in follower counts, where some profiles had values vastly exceeding the norm. To handle these anomalies, we employed an Interquartile Range (IQR) method, which effectively identifies and removes extreme values without compromising the dataset's overall distribution. A systematic approach for handling infinite values in numeric columns was implemented, replacing such values with numpy.nan before applying median imputation.

We implemented standardization through scikit-learn's StandardScaler, which transformed our numeric features to zero mean and unit variance. This scaling was applied to features including age, beauty, number of previous positions, total experience, average time in previous position, current position length, and career progression rate. The transformation followed the formula:

\[z = \frac{x - \mu}{\sigma}\]

where $x$ represents the original feature value, $\mu$ is the mean, and $\sigma$ is the standard deviation. This standardization proved crucial for maintaining feature comparability and ensuring optimal performance of our machine learning algorithms.

For categorical variables, we employed scikit-learn's OneHotEncoder with the parameter drop='first' to avoid multicollinearity. This transformation converted categorical variables like industry and nationality into 19 distinct binary features. The encoding process maintained feature sparsity while eliminating the ordinal relationships that might be incorrectly implied by label encoding.

To address class imbalance in our dataset, we implemented a two-stage sampling approach. First, we applied RandomUnderSampler to reduce the majority class, followed by RandomOverSampler to balance the minority classes. This sequential approach was particularly effective for gender and ethnicity distributions, where initial ratios showed significant imbalances (76.2\% male, 76.4\% White). The sampling was configured with:
\[sampling_strategy='auto'
random_state=42\]
Feature engineering expanded our analysis capabilities through calculated metrics. Most notably, we developed a career progression rate metric that considered both historical and current career trajectories:

\textbf{\begin{equation*}
\begin{array}{l}
\mathit{career\_progression\_rate} = \\[10pt]
\frac{\mathit{no\_of\_promotions}}{
  \begin{array}{l}
    (\mathit{avg\_previous\_position\_length} \times \\
    \mathit{no\_of\_previous\_positions} + \\
    \mathit{current\_position\_length})
  \end{array}
}
\end{array}
\end{equation*}}

This formulation provides a normalized measure of career advancement speed, accounting for both the frequency of promotions and the time taken to achieve them. The \(career progression rate\) calculation initially produced 13 missing values due to division operations, which were addressed through median imputation to maintain consistency.
The pre-processing decisions were validated through cross-validation, ensuring our handling of missing and infinite values did not introduce artifacts into the subsequent analysis while maintaining the interpretability of our features. This comprehensive approach to data preparation provided a robust foundation for our machine learning models while preserving the essential characteristics of the original dataset.

\subsection{Data Visualization and Exploratory Analysis}
Our initial data exploration revealed several key insights through visualization techniques. We employed a combination of distribution plots, correlation analyses, and outlier detection to better understand the relationships between different profile characteristics and success metrics.
\begin{figure}
    \centering
    \includegraphics[width=1.00\linewidth]{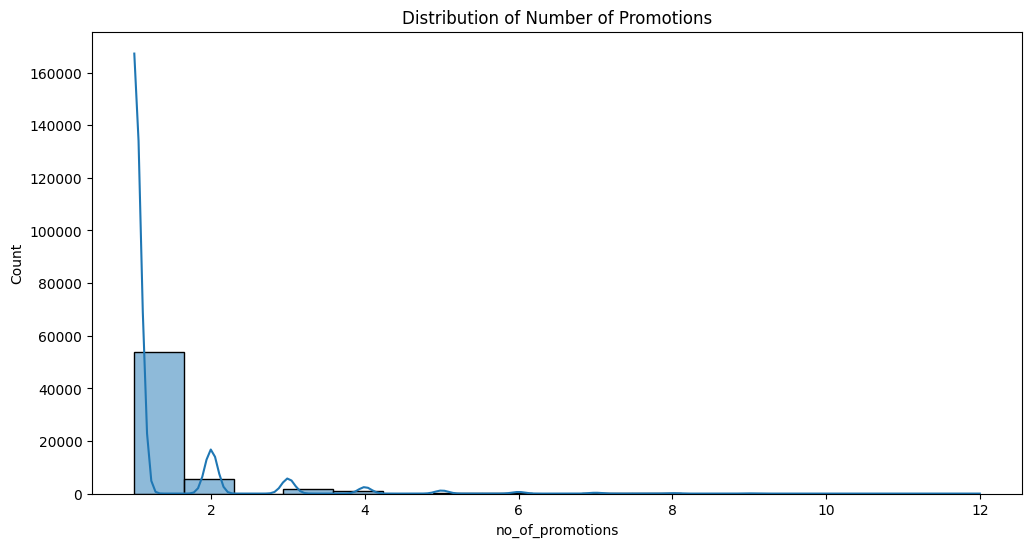}
    \caption{Frequency distribution of promotions among LinkedIn professionals (N=62,650)}
    \Description{A histogram showing the frequency distribution of promotions among LinkedIn professionals.}
    \label{fig:distribution}
\end{figure}

\begin{figure}
    \centering
    \includegraphics[width=1.00\linewidth]{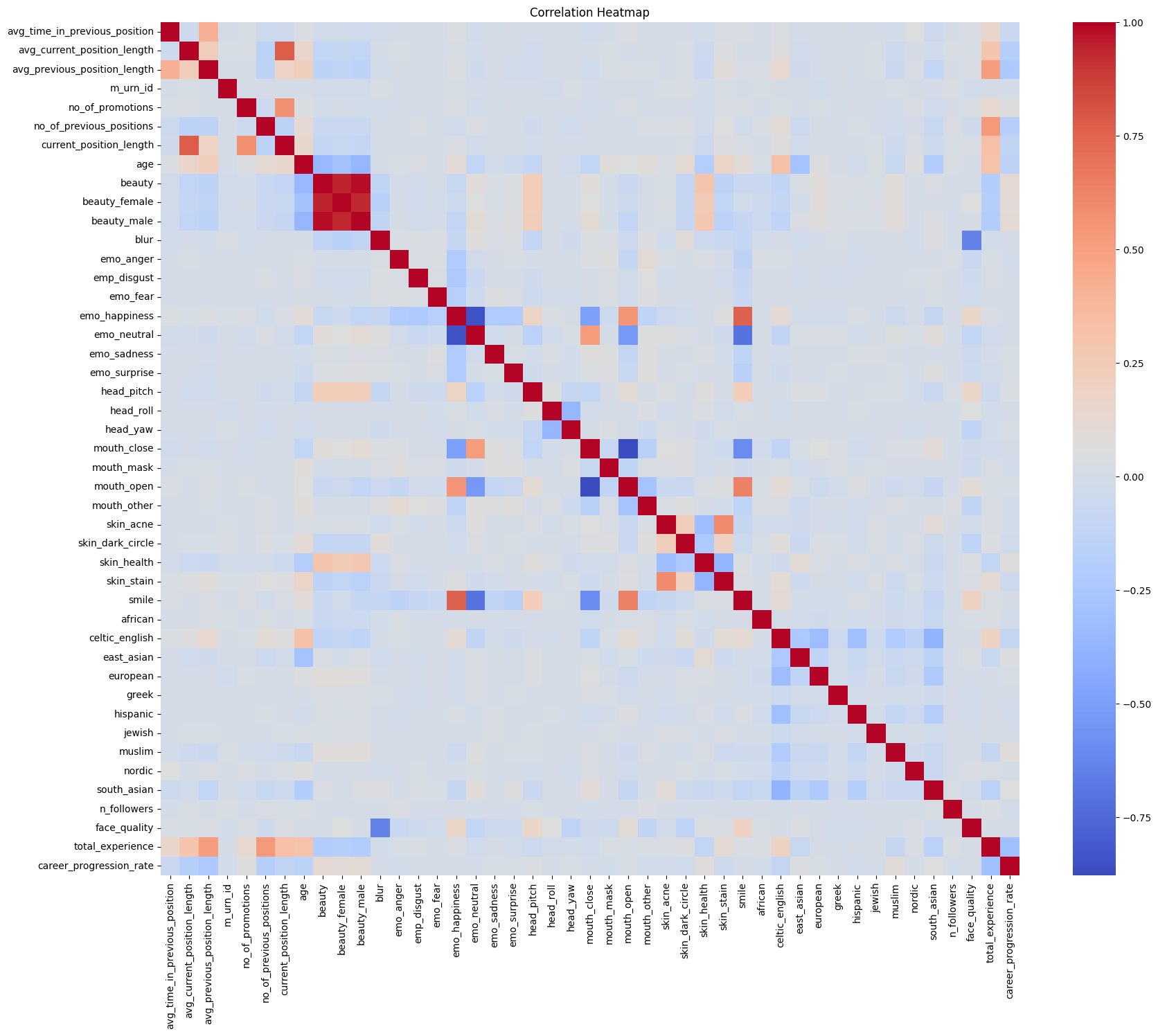}
    \caption{Correlation heatmap of LinkedIn profile features (N=62,650)}
    \label{fig:heatmap}
\end{figure}

\begin{figure} [!t]
    \centering
    \includegraphics[width=1.00\linewidth]{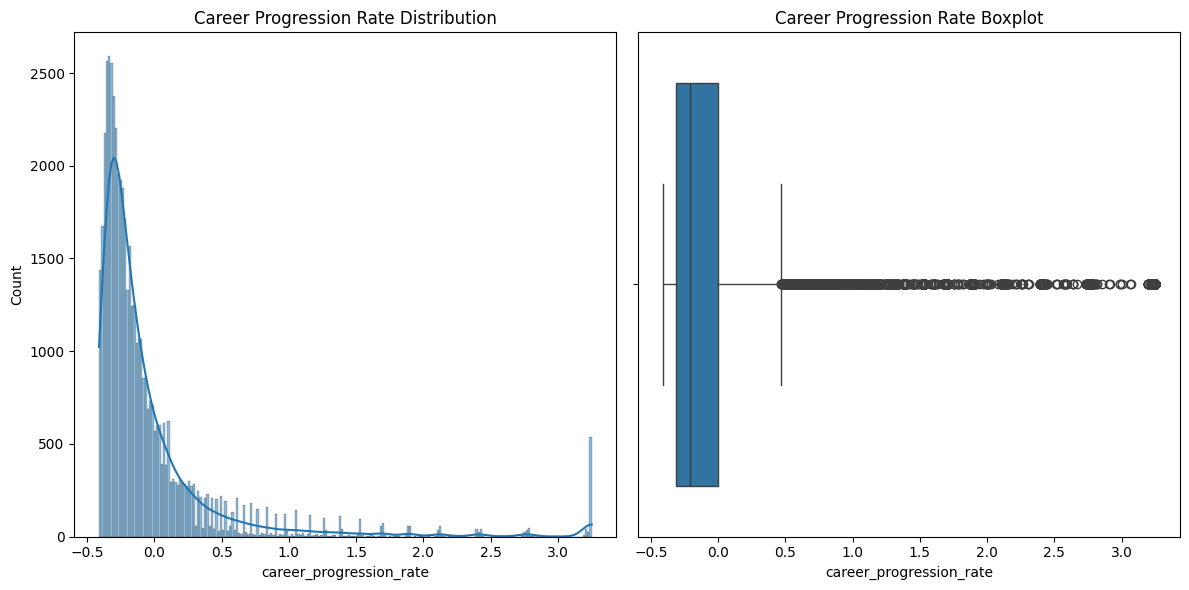}
    \caption{Residuals vs Predicted Followers boxplot analysis (N=62,650)}
    \label{fig:boxplot}
\end{figure}

\subsubsection{Distribution Analysis}
The career progression rate distribution Figure \ref{fig:distribution} showed a distinct right-skewed pattern, with most professionals having moderate progression rates (between -0.5 and 0.5) and a notable tail extending towards higher rates. This indicates that while rapid career advancement is possible, it's relatively rare in our dataset. The distribution statistics revealed a mean of -0.000050, standard deviation of 0.999895, with values ranging from -0.450598 to 52.019582.

\subsubsection{Correlation Analysis}
The correlation heatmap \ref{fig:heatmap} revealed several significant relationships among profile characteristics. Career progression rate demonstrated a strong positive correlation (0.472032) with number of promotions, while current position length showed substantial correlation (0.313682) with total experience. Beauty score emerged as a significant predictor of follower count, with a correlation coefficient of 0.352549. Age and total experience showed moderate correlation (0.135577), as did current position length with career progression (0.282958). Notably, beauty score and age showed minimal correlation with promotions (0.001329 and 0.001027 respectively).

\subsubsection{Outlier Analysis}
Our boxplot analysis \ref{fig:boxplot} identified several notable patterns in the data distribution. Follower counts showed significant right-skew, with a median around 2,000 followers but outliers extending beyond 400,000 followers. The interquartile range indicated that 75\% of profiles had fewer than 10,000 followers.

Career progression rates clustered primarily between -0.313689 and -0.000598, with notable outliers reaching 52.019582. The distribution maintained relative symmetry around the median, though with a pronounced right tail. Position length analysis revealed most roles lasting between one and three years, with some significant outliers extending beyond ten years. The data showed a clear lower bound at zero years, as expected.

These visualization findings significantly influenced our subsequent modeling decisions, particularly regarding outlier handling and feature selection. The presence of extreme outliers, especially in follower counts, guided our choice of robust modeling techniques and informed our feature engineering approach.

\subsection{Model Training}
Our analysis employed four distinct machine learning models, each configured with specific hyperparameters optimized for our dataset.
Known for its robustness and ability to handle non-linear relationships, Random Forest served as a reliable starting point. We implemented an ensemble of 100 trees with a fixed random state of 42 to ensure reproducibility. This configuration provided a balance between computational efficiency and model performance.

Next, we introduced Gradient Boosting, which was selected for its ability to learn from residual errors and refine predictions iteratively. The model was configured with a learning rate of 0.2 and a maximum depth of 3, utilizing 300 estimators. These parameters were selected through cross-validation to prevent overfitting while maintaining strong predictive performance. The relatively high learning rate combined with a controlled tree depth allowed the model to capture complex patterns while remaining generalizable.

For the XGBoost implementation, we utilized 200 estimators with a more conservative learning rate of 0.1 and a maximum depth of 5. This configuration allowed for gradual improvement in prediction accuracy while managing the risk of overfitting. The increased maximum depth, compared to the Gradient Boosting model, enabled XGBoost to capture more complex feature interactions.

The CatBoost model was configured with 200 iterations, matching the scale of our XGBoost implementation. We maintained a learning rate of 0.1 and set the tree depth to 5, allowing for effective handling of categorical variables while maintaining model stability. CatBoost's internal handling of categorical features made it particularly suitable for our dataset's mixed numerical and categorical nature.

These parameter selections were determined through extensive experimentation and cross-validation, with final configurations chosen based on their performance on our validation set. All models maintained consistent random seeds to ensure reproducibility of results and fair comparison across different approaches.
 \section{Results and Analysis}
The performance of each model was evaluated through a comprehensive set of metrics including cross-validation scores, mean squared error (MSE), and residual analysis. 

Tables \ref{table:model_performance_promotions} and \ref{table:model_performance_followers} presents the detailed performance metrics across all models for both promotion and follower count predictions.

\begin{table}[ht]
\caption{Model Performance Comparison for Promotions Prediction}
\centering
\begin{tabular}{|l|c|c|c|}
\hline
\textbf{Model} & \textbf{MSE} & \textbf{R² Score} & \textbf{Cross-validation R²} \\
& & & (Mean ± Std) \\
\hline
Random Forest & 0.0025 & 0.9959 & 0.9930 ± 0.0018 \\ \hline
Gradient Boosting & 0.0144 & 0.9763 & 0.9706 ± 0.0027 \\ \hline
XGBoost & 0.0025 & 0.9959 & 0.9936 ± 0.0009 \\ \hline
CatBoost & 0.0019 & 0.9968 & 0.9950 ± 0.0008 \\ 
\hline
\end{tabular}
\label{table:model_performance_promotions}
\end{table}

\begin{table}[ht]
\caption{Model Performance Comparison for Followers Prediction}
\centering
\begin{tabular}{|l|c|c|c|}
\hline
\textbf{Model} & \textbf{MSE} & \textbf{R² Score} & \textbf{Cross-validation R²} \\
& & & (Mean ± Std) \\
\hline
Random Forest & 15,822,758 & 0.3874 & 0.0654 ± 0.0897 \\ \hline
Gradient Boosting & 5,929,601 & 0.7704 & 0.6291 ± 0.0897 \\ \hline
XGBoost & 30,671,037 & -0.1875 & -0.0261 ± 0.1559 \\ \hline
CatBoost & 23,102,439 & 0.1055 & 0.0393 ± 0.1889 \\
\hline
\end{tabular}
\label{table:model_performance_followers}
\end{table}

Residual analysis revealed distinct patterns across different prediction tasks. For promotions prediction, all models demonstrated relatively uniform residual distributions, with CatBoost showing the most consistent performance (residual standard deviation = 0.0436). The residuals followed an approximately normal distribution, indicating well-calibrated predictions across the range of values.
In contrast, follower count predictions showed heteroscedastic residual patterns, with error variance increasing for profiles with higher follower counts. Gradient Boosting demonstrated the most stable residual distribution among all models, with a mean residual of 6.107707 and standard deviation of 1367.559183. The residual distribution exhibited the following quartile characteristics:

\[Q_1 (25th percentile): -613.502577\]
\[Q_2 (Median): -272.152041\]
\[Q_3 (75th percentile): 206.021182\]

Cross-validation results highlighted the stability of promotion predictions across all models, with consistently high R² scores and low standard deviations. However, follower count predictions showed substantial variation across folds, particularly for XGBoost and CatBoost, suggesting these models may be more sensitive to the specific characteristics of the training data.
The significant difference in prediction accuracy between promotions and followers (mean R² of 0.9912 vs 0.2690 across models) indicates that career advancement follows more structured patterns than social network growth. This disparity suggests that while professional progression can be effectively modeled using traditional career metrics, follower acquisition may depend on additional factors not captured in our current feature set.

\subsection{Promotions Prediction Performance}
Promotions prediction showed exceptional consistency across all models, with R² values exceeding 0.97. This high level of accuracy suggests that promotions are highly structured and strongly correlated with the features in our dataset. CatBoost had a slight edge over the other models, owing to its superior handling of categorical data. The consistency in performance across all models underscores the robustness of the relationship between features like career progression rate and promotions.

\begin{figure}[ht]
  \centering
  \includegraphics[width=\linewidth]{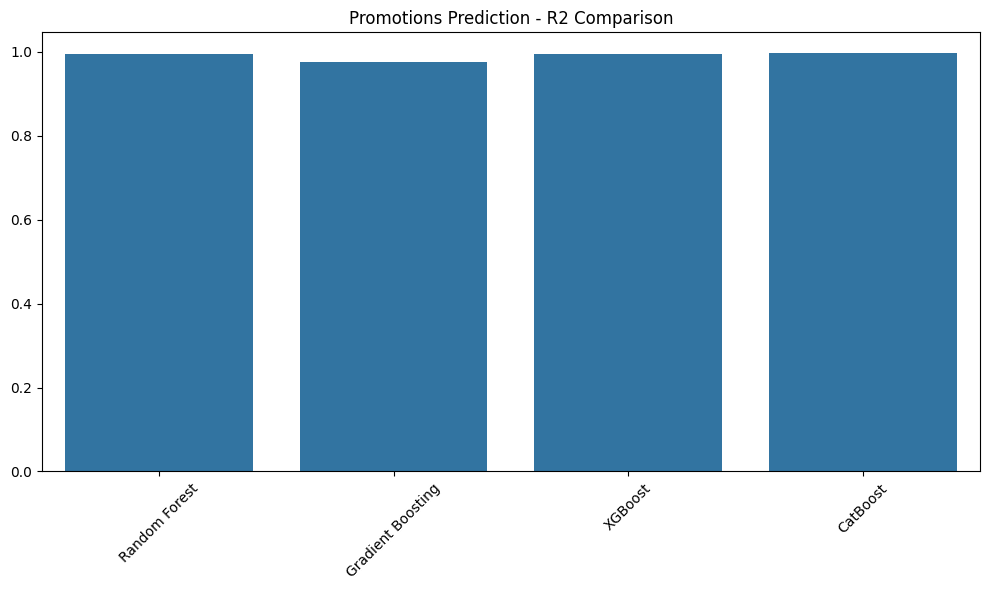}
   \caption{R² Comparison for Promotions Prediction Across models}
\end{figure}
\begin{figure}[h]
  \centering
  \includegraphics[width=\linewidth]{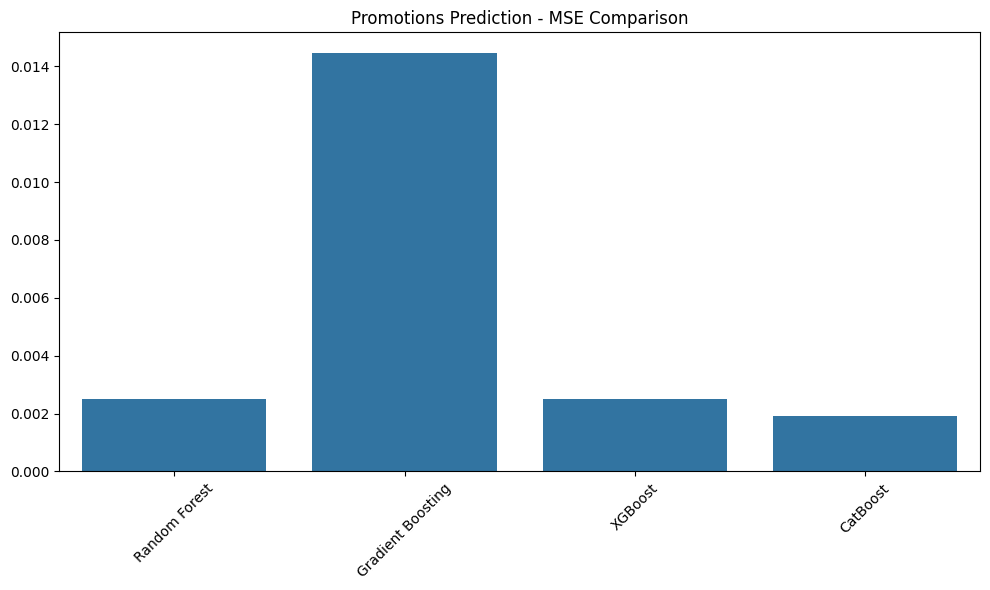}
   \caption{MSE Comparison for Promotions Prediction Across models}
\end{figure}

\subsection{Followers Prediction Performance}
In contrast to promotions, predicting follower count revealed a stark variation in model performance. Gradient Boosting significantly outperformed other models, achieving an R² of 0.7704, demonstrating its ability to capture the complexity of follower growth. XGBoost, however, recorded a negative R², indicating it performed worse than a mean prediction. These results suggest that follower growth is influenced by complex and non-linear patterns that are not as easily captured by all models.

\begin{figure}[ht]
  \centering
  \includegraphics[width=\linewidth]{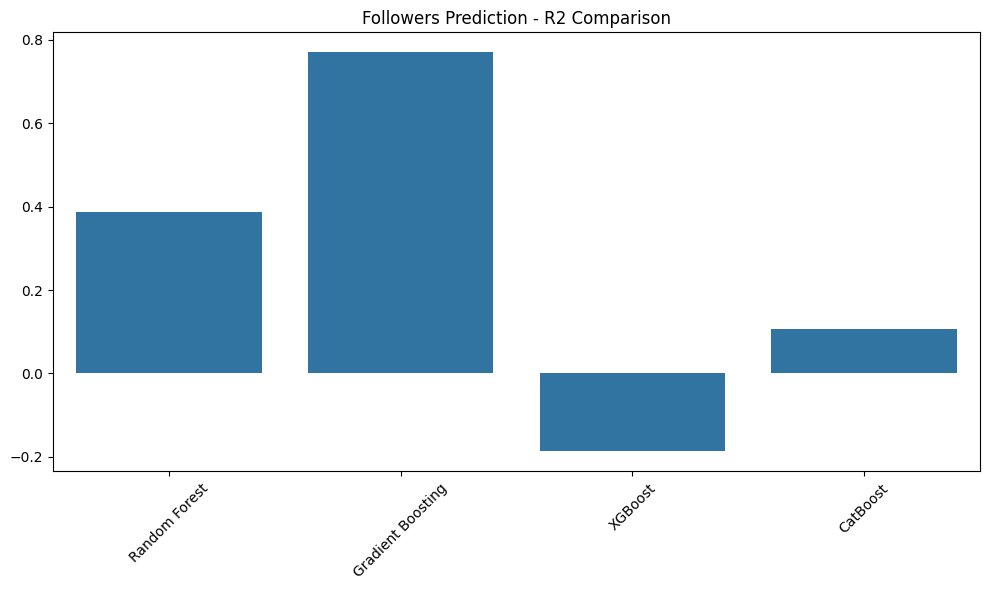}
   \caption{R² Comparison for Followers Prediction Across models}
\end{figure}
\begin{figure}[h]
  \centering
  \includegraphics[width=\linewidth]{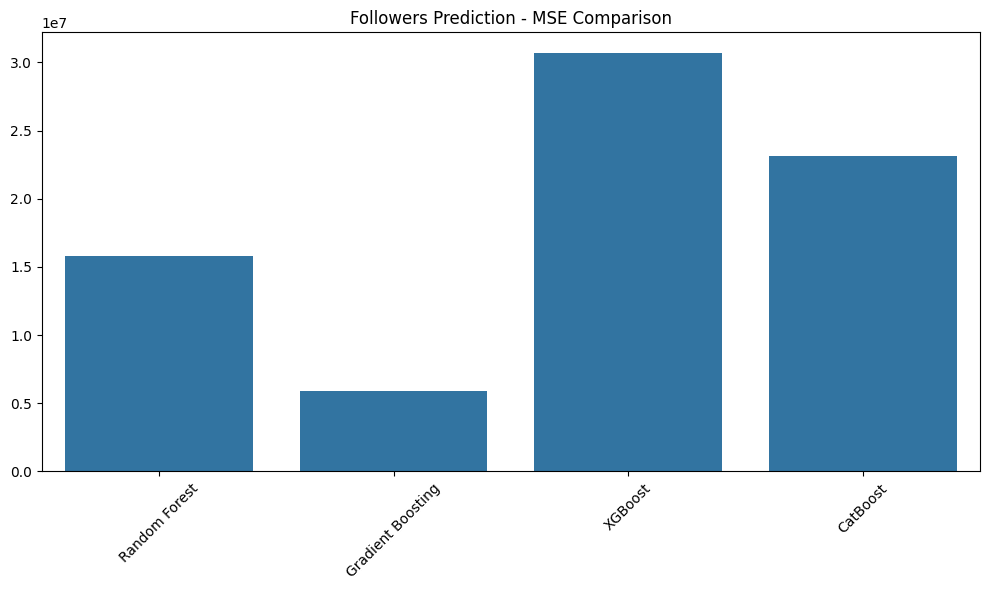}
   \caption{MSE Comparison for Followers Prediction Across models}
\end{figure}

\subsection{Feature Importance}
Feature importance analysis across different models revealed consistent patterns while highlighting some model-specific variations in feature ranking. Tables \ref{table:feature_importance_promotions} and \ref{table:feature_importance_followers} presents the importance scores of top features for promotion prediction across all models.

The analysis revealed several compelling insights about feature importance across our models. For promotion prediction, \(career_progression_rate\) consistently emerged as the most influential feature across all models, explaining approximately 47-48\% of the variance in predictions. This dominance was followed by \(current_position_length\) accounting for 30-31\% of importance, and \(total_experience\) contributing approximately 21\%. These consistent rankings suggest a robust relationship between these career metrics and professional advancement.

In contrast, follower count prediction exhibited more diverse feature importance patterns. While beauty score dominated across all models, its relative importance showed significant variation, ranging from 35\% in Random Forest to a striking 81\% in CatBoost. Similarly, \(current_position_length\) demonstrated substantial variability in its predictive power, accounting for 28\% of importance in Random Forest but dropping to merely 5.7\% in CatBoost. This variability suggests that different models capture distinct aspects of social network growth on the platform.

Demographic features presented an interesting dichotomy in their predictive significance. Age, for instance, showed consistently low importance for promotion prediction, accounting for less than 0.1\% of feature importance. However, it demonstrated moderate importance for follower prediction, contributing between 11-13\% across most models. This contrast suggests that while demographic factors play a minimal role in career advancement, they may influence social network development on LinkedIn.

The consistency in feature importance rankings for promotion prediction suggests robust relationships between career metrics and advancement opportunities. Conversely, the variation in feature importance for follower prediction indicates that different models capture distinct aspects of social network growth, potentially explaining the differences in prediction accuracy across models.
Notably, categorical features such as nationality and ethnicity showed minimal importance across all models and prediction tasks, with importance scores consistently below 0.1\%. This suggests that career advancement and social network growth in our dataset are primarily driven by professional metrics and profile characteristics rather than demographic factors.

\begin{table}[ht]
\caption{Feature Importance for Promotions Prediction}
\small
\centering
\begin{tabular}{l|r|r|r|r}
\hline
\textbf{Feature} & \textbf{RF} & \textbf{GB} & \textbf{XGB} & \textbf{CB} \\
\hline
career\_prog\_rate & 0.4720 & 0.4809 & 0.4720 & 0.4809 \\ \hline
current\_pos\_len & 0.3137 & 0.3049 & 0.3137 & 0.3049 \\ \hline
total\_experience & 0.2119 & 0.2112 & 0.2119 & 0.2112 \\ \hline
beauty & 0.0013 & 0.0009 & 0.0013 & 0.0009 \\ \hline
age & 0.0010 & 0.0007 & 0.0010 & 0.0007 \\
\hline
\end{tabular}
\begin{flushleft}\small RF: Random Forest, GB: Gradient Boosting, XGB: XGBoost, CB: CatBoost\end{flushleft}
\label{table:feature_importance_promotions}
\end{table}

\begin{table}[ht]
\caption{Feature Importance for Followers Prediction}
\small
\centering
\begin{tabular}{l|r|r|r|r}
\hline
\textbf{Feature} & \textbf{RF} & \textbf{GB} & \textbf{XGB} & \textbf{CB} \\
\hline
beauty & 0.3525 & 0.3382 & 0.3525 & 0.8094 \\ \hline
current\_pos\_len & 0.2830 & 0.1873 & 0.2830 & 0.0573 \\ \hline
age & 0.1356 & 0.1153 & 0.1356 & 0.0155 \\ \hline
career\_prog\_rate & 0.1074 & 0.0860 & 0.1074 & 0.0090 \\ \hline
total\_experience & 0.1215 & 0.0843 & 0.1215 & 0.0267 \\
\hline
\end{tabular}
\begin{flushleft}\small RF: Random Forest, GB: Gradient Boosting, XGB: XGBoost, CB: CatBoost\end{flushleft}
\label{table:feature_importance_followers}
\end{table}

\begin{figure}[ht]
  \centering
  \includegraphics[width=\linewidth]{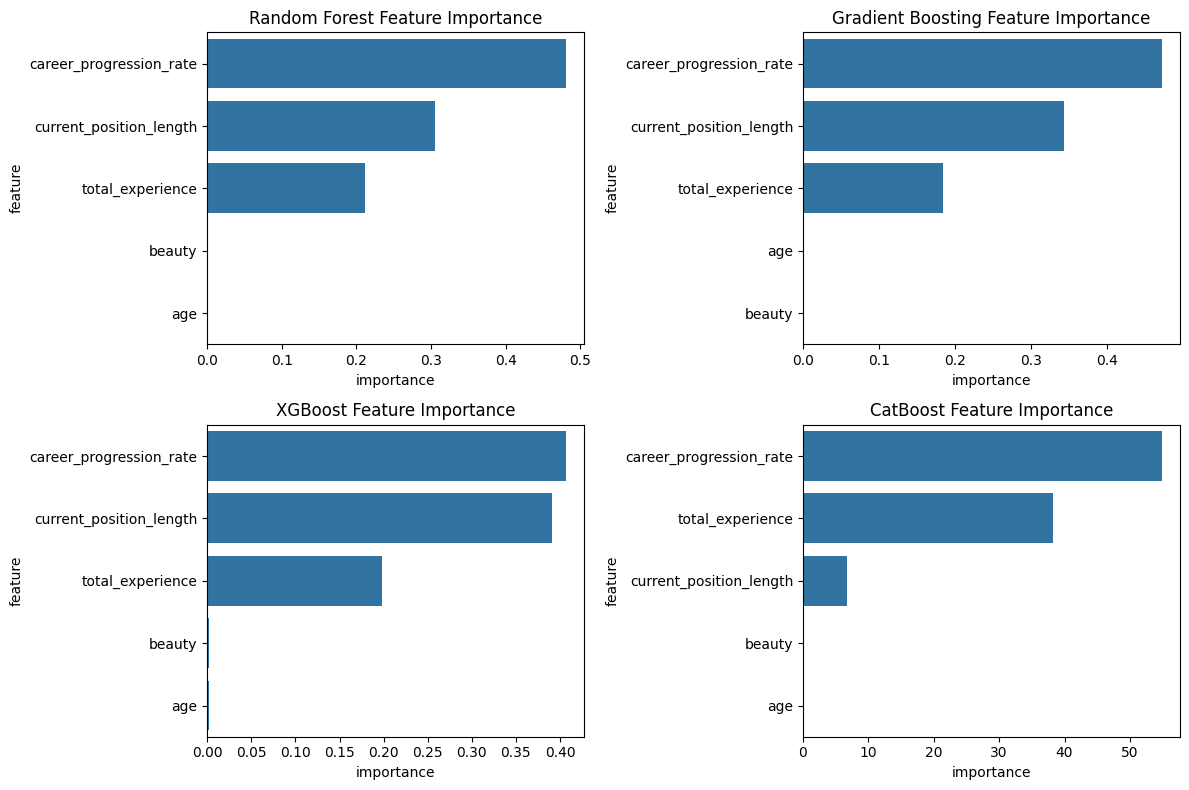}
   \caption{Feature Importance for Followers and Promotions}
\end{figure}

\subsection{Error Analysis}
\begin{figure}[ht]
  \centering
  \includegraphics[width=\linewidth]{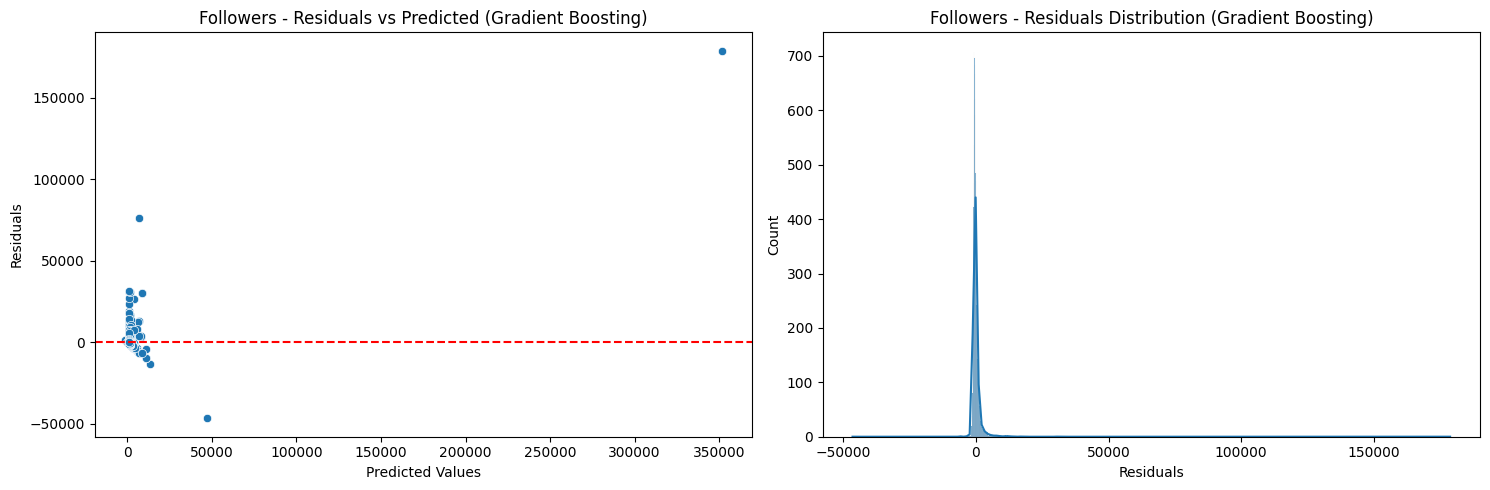}
   \caption{Residuals vs Predicted (Gradient-Boosting}
   \label{fig:residuals_gradient}
\end{figure}

Residual analysis revealed distinct patterns in prediction accuracy for both follower count and promotion predictions. Figure \ref{fig:residuals_gradient} shows the error analysis for follower predictions using Gradient Boosting, where the residuals versus predicted values plot exhibits a clear heteroscedastic pattern. The scatter plot demonstrates increasing error variance as predicted follower counts grow, with residuals ranging from approximately -50,000 to over 150,000 for higher predictions. The residuals distribution shows a sharp peak near zero with long tails, particularly in the positive direction, indicating a tendency for some significant overestimations in follower predictions.

\begin{figure}[ht]
  \centering
  \includegraphics[width=\linewidth]{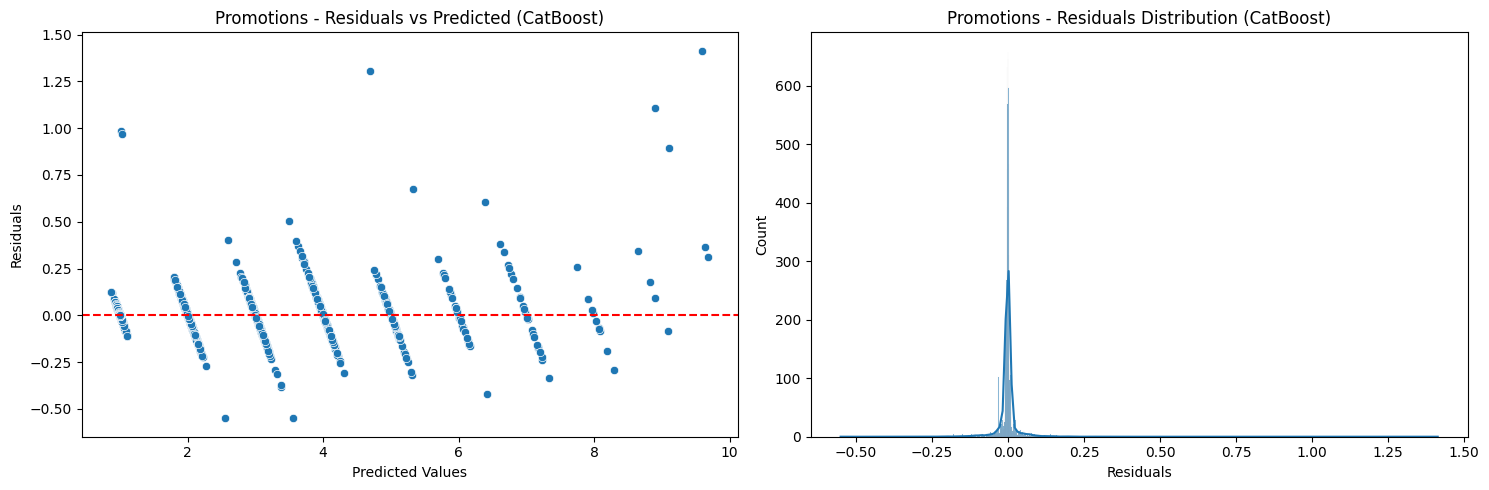}
   \caption{Residuals vs Predicted (CatBoost}
   \label{fig:catboost_gradient}
\end{figure}

Figure \ref{fig:catboost_gradient} presents the error analysis for promotion predictions using CatBoost, revealing markedly different patterns. The residuals versus predicted values plot shows a more structured pattern with distinct vertical bands, suggesting quantized predictions typical of promotion counts. The residuals remain largely bounded between -0.5 and 1.5, with most errors clustered tightly around zero. The residuals distribution demonstrates a highly concentrated bell curve around zero, indicating the model's strong predictive performance for promotions. This tight distribution of residuals, combined with the symmetric pattern around zero, suggests well-calibrated promotion predictions across different ranges of the target variable.

The contrasting error patterns between follower and promotion predictions highlight the differing challenges in predicting these two metrics. While promotion predictions show consistent accuracy across the range of predicted values, follower predictions become increasingly uncertain for profiles with larger follower counts, suggesting that additional features or model refinements might be needed to better capture the factors influencing high follower counts.

\section{Conclusion}
Our analysis highlights significant differences in the predictability of promotions and follower growth on LinkedIn, offering valuable insights into professional success metrics. For promotions, we found a highly structured and predictable relationship with the features in our dataset. All models performed exceptionally well, achieving R² values above 0.97, with CatBoost offering a slight edge due to its superior handling of categorical data. The consistency across models indicates that the key factors driving promotions, such as career progression rate and position tenure, are well-represented in the dataset. This makes any model a viable choice for predicting promotions, providing a reliable and robust prediction system.

In contrast, follower growth proved to be more complex, with Gradient Boosting emerging as the most effective model. Its ability to handle non-linear relationships allowed it to achieve an R² of 0.7704, outperforming other models. However, the variability in model performance and higher error rates underscore the dynamic and multifaceted nature of social proof metrics. Features such as beauty score played a significant role in follower predictions, reflecting the importance of profile aesthetics on LinkedIn.

To improve follower growth predictions, we recommend focusing on advanced feature engineering. Incorporating additional features, such as metrics for network interactions and career trajectory patterns, could better capture the dynamics influencing follower growth. Developing ensemble approaches may also enhance the performance of future models.

In conclusion, our analysis demonstrates that promotions can be predicted with high accuracy using current profile characteristics, while follower growth requires further investigation due to its complexity. We recommend CatBoost as the preferred model for promotion predictions and Gradient Boosting for follower predictions. This study provides actionable insights for LinkedIn users seeking to optimize their profiles, as well as organizations looking to evaluate professional success through data-driven methods.

\printbibliography

\end{document}